\documentclass[cameraready]{Interspeech}

\title{Gated Multi-Graph Fusion via Graph Attention Networks for Alzheimer’s Disease Detection}

\author[affiliation={1}]{Jinyu}{Li}
\author[affiliation={1,2}]{Xiao}{Wei}
\author[affiliation={1,2}]{Bin}{Wen}
\author[affiliation={2}]{Kai}{Li}
\author[affiliation={3}]{Yuqin}{Lin}
\author[affiliation={1}]{Xiaobao}{Wang}
\authorbreak
\author[affiliation={1,4}, correspondingauthor]{Longbiao}{Wang}
\author[affiliation={2}, correspondingauthor]{Jianwu}{Dang}

\address{
    $^1$ School of Future Technology, Tianjin University, Tianjin, China \\
    $^2$ Shenzhen Institutes of Advanced Technology, Chinese Academy of Sciences, Shenzhen, China \\
    $^3$ College of Computer and Data Science, Fuzhou University, Fuzhou, China \\
    $^4$ Huiyan Technology (Tianjin) Co., Ltd, Tianjin, China
}

\email{lijinyu536@tju.edu.cn, longbiao\_wang@tju.edu.cn}


\usepackage{comment}
\usepackage{amsmath}
\usepackage{amssymb}
\usepackage{multirow}
\usepackage{booktabs}  
\usepackage{bm}        
\newcommand{\concat}{\mathbin{\Vert}}

\begin{document}

\maketitle

\begin{abstract}
    Spontaneous speech is a vital non-invasive biomarker for Alzheimer’s Disease (AD), yet many systems overlook non-linear structural disruptions and clinical heterogeneity in pathological language. We propose a Multi-View Gated Graph Attention Network that transcribes audio via Automatic Speech Recognition (ASR) to construct semantic, dependency, and co-occurrence graphs, characterizing speech through a "content–structure–flow" framework. Notably, the co-occurrence graph leverages Pointwise Mutual Information (PMI) from a normative corpus to quantify narrative logic and linguistic deviation. To address symptomatic diversity, an adaptive gated fusion mechanism dynamically integrates these views. Evaluated on the ADReSSo dataset, our model achieves 90.00\% accuracy. Ablation results confirm that the PMI-based graph and heterogeneity-aware gating are essential for robust classification across diverse clinical populations. Our source code is publicly available at https://github.com/opeacc/AD.
\end{abstract}

\section{Introduction}
    Dementia, primarily AD, represents an escalating global health crisis hallmarked by progressive cognitive decline that manifests early in spontaneous speech \cite{lindsay_language_2021,gumus_linguistic_2024}. The "Cookie Theft" picture description task is extensively recognized for its clinical utility in eliciting these critical linguistic markers in a controlled environment \cite{qi_noninvasive_2023}. Initial efforts in speech-based dementia detection predominantly relied on traditional machine learning paradigms \cite{qiao_alzheimer8217s_2021,kurdi}. These methodologies utilized manual feature engineering to extract handcrafted acoustic parameters—such as pitch and jitter—alongside lexical diversity metrics, which were subsequently classified using algorithms like Support Vector Machines (SVM) or Random Forests \cite{syed_tackling_2021,shankar_systematic_2025}. While these models offered high interpretability \cite{calza_linguistic_2021}, they were constrained by their dependence on expert-defined features and their inherent inability to capture the latent, non-linear dependencies found in spontaneous discourse. The subsequent evolution of deep learning, specifically the application of Large Language Models (LLMs) such as BERT, has substantially improved diagnostic performance \cite{balagopalan_comparing_2021,wei_breaking_2026}. Nonetheless, these models primarily learn semantic representations implicitly \cite{balagopalan_bert_2020,ajroudi_exploring_2024,chlasta_enhancing_2025} and often fail to sufficiently characterize the complex, multi-dimensional structural degradation characteristic of pathological speech. Recently, researchers have explored the use of Graph Neural Networks (GNNs) to model the intricate topological structure of language, yielding promising results \cite{cai_exploring_2023,hallani_graph_2025}. However, current research in this domain remains insufficient in its modeling of linguistic features. Furthermore, existing frameworks typically rely on simplistic fusion strategies (e.g., concatenation or fixed-weight summation), limiting the model's adaptability to diverse symptomatic manifestations across the clinical spectrum.
    
    In this paper, we propose a multi-view graph learning framework that addresses two critical gaps in current research through the following innovations: First, we introduce a "content–structure–flow" trinity for comprehensive linguistic modeling. Traditional methods often overlook the "flow" of discourse—the logical progression of speech. While semantic graphs capture "content" and dependency graphs capture "structure," we specifically design a Co-occurrence Graph based on PMI derived from healthy normative data. This graph uniquely reflects a subject's ability to describe local events. In patients with AD, speech often exhibits disordered sequences, logical jumps, and repetitive loops \cite{burke_comparing_2023}. By integrating this co-occurrence perspective, our model provides a full-spectrum characterization of linguistic patterns, capturing not just what is said, but how the narrative logic unfolds. Second, we propose a Gated Fusion Mechanism to address the clinical heterogeneity of AD. Clinical observations demonstrate that dementia symptoms are not monolithic: some patients exhibit "syntactic collapse" (simplified grammar with preserved vocabulary), while others suffer from "semantic void" (fluent but meaningless speech) \cite{fraser_linguistic_2015}. Our gated network adaptively assigns weights to the semantic, dependency, and co-occurrence representations on a per-sample basis. This allows the framework to capture sample-specific features, dynamically focusing on the most discriminative linguistic markers for each individual.

    In summary, our contributions are as follows:
    \begin{enumerate}
    
    \item Discourse Flow Analysis: We utilize a PMI-based co-occurrence graph to quantify the deviation of event description logic from healthy norms.

    \item A Holistic Multi-Graph Framework: We integrate semantic, dependency, and co-occurrence graphs to model the "content–structure–flow" of spontaneous speech.
    
    \item Heterogeneity-aware Fusion: We implemented a gated fusion mechanism that accounts for the diversity of AD symptoms, enhancing the model's ability to adapt to varying symptomatic presentations.
    \end{enumerate}

\begin{figure*}[t]  
    \centering
    \includegraphics[width=0.9\textwidth]{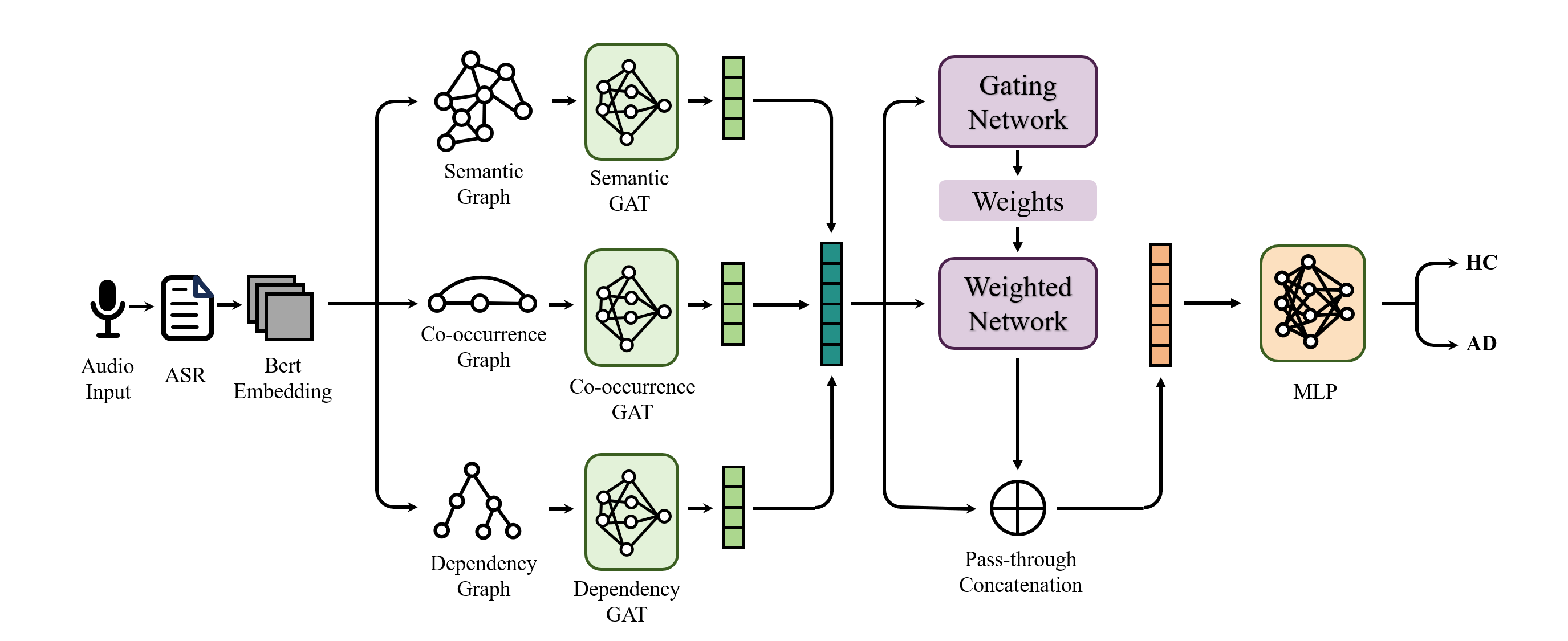}
    \caption{\textbf{Overview of the proposed Multi-View Gated Graph Attention Network framework for dementia detection.} (1) Audio is transcribed via ASR and converted into word-level embeddings. (2) Semantic, co-occurrence, and dependency graphs are constructed to capture multi-scale linguistic patterns. (3) Topological features are extracted by GAT layers and aggregated via global pooling. (4) These multi-view features are adaptively fused by a gating network. (5) The fused features are fed into an MLP for final classification.}
    \label{fig:double}
\end{figure*}

\section{Multi-View Gated GAT}
We propose a multi-dimensional graph learning framework, as illustrated in Figure~\ref{fig:double}, designed to capture the "content–structure–flow" of spontaneous speech. This architecture consists of five main modules as detailed below.

\subsection{Speech Transcription and Node Embedding}

\textbf{Automatic Speech Recognition (ASR):} Given the raw audio signal of a subject describing the "Cookie Theft" picture, we utilize Whisper \cite{pmlr-v202-radford23a} for transcription. Whisper’s large-scale weak supervision training makes it exceptionally robust to the disfluencies common in dementia-impacted speech. The output is a sequence of $N$ words: $S = \{w_1, w_2, \dots, w_N\}$.

\noindent
\textbf{Node Feature Initialization:} To represent each word in a high-dimensional semantic space, we use a pre-trained BERT-base \cite{devlin_bert_2019} model. Since BERT employs a WordPiece tokenizer, a word $w_i$ may be split into $k$ sub-word tokens $\{t_{i,1}, t_{i,2}, \dots, t_{i,k}\}$. To maintain a 1-to-1 mapping between the transcript and graph nodes, we define the node representation $\mathbf{x}_i \in \mathbb{R}^{768}$ as the mean of its constituent token embeddings:
\begin{equation}
\mathbf{x}_i = \text{MeanPooling}(\text{BERT}(t_{i,1}), \dots, \text{BERT}(t_{i,k})).
\end{equation}
The final node feature matrix for a transcript is $\mathbf{X} \in \mathbb{R}^{N \times 768}$.

\subsection{Multi-View Graph Construction}
We define three graphs $\mathcal{G}_{sem}$, $\mathcal{G}_{syn}$, and $\mathcal{G}_{co}$ sharing the same vertex set $\mathcal{V}$, where $|\mathcal{V}| = N$.

\subsubsection{Semantic Graph: Content Representation}
This graph models the conceptual density and semantic relationships. We compute the cosine similarity between every pair of word embeddings $\mathbf{x}_i$ and $\mathbf{x}_j$ \cite{chen_iterative_2020}. An edge $e_{ij} \in \mathcal{E}_{sem}$ is created if the similarity exceeds a predefined threshold $\tau_{s}$:
\begin{equation}
\operatorname{Sim}(\mathbf{x}_i, \mathbf{x}_j) = 
\frac{\mathbf{x}_i^\top \mathbf{x}_j}
     {\lVert \mathbf{x}_i \rVert \, \lVert \mathbf{x}_j \rVert} 
> \tau_{s}.
\end{equation}
This captures the global semantic structural characteristics of the subject.

\subsubsection{Dependency Graph: Structural Integrity}
To capture grammatical decay (e.g., loss of complex clauses), we use spaCy \cite{honnibal2017spacy} to perform dependency parsing. An edge $e_{ij} \in \mathcal{E}_{syn}$ is established if there is a direct syntactic dependency (e.g., nominal subject, direct object, modifier) between $w_i$ and $w_j$. This graph reflects the structural complexity of the subject’s language \cite{ivanova_defying_2023,lian_dependency_2025}.

\subsubsection{Co-occurrence Graph: Discourse Flow via Normative PMI}
To capture the "flow" dimension of the "content-structure-flow" trinity, we design a co-occurrence graph that quantifies the logical progression and temporal coherence of speech. Traditional sequential models often fail to detect the subtle "logical jumps" or repetitive loops characteristic of AD-impacted discourse. Our approach addresses this by measuring how much a subject's word associations deviate from established healthy linguistic norms.

We first construct a Normative Corpus $\mathcal{D}_{hc}$ using only transcripts from healthy controls. This corpus serves as a linguistic reference baseline for characterizing natural word associations within the context of the 'Cookie Theft' task. We calculate the Pointwise Mutual Information (PMI) \cite{church_word_1990,yao_graph_2018} for all word pairs $(w_i, w_j)$ that appear within a sliding window of size $n$ across $\mathcal{D}_{hc}$:
\begin{equation}
\text{PMI}(w_i, w_j) = \log \frac{P(w_i, w_j)}{P(w_i)P(w_j)},
\end{equation}
where $P(w_i)$ and $P(w_j)$ are the individual unigram probabilities, and $P(w_i, w_j)$ is the joint probability of co-occurrence within the window. A high PMI indicates a strong, statistically significant association between words (e.g., "sink" and "overflowing") that reflects healthy narrative logic.

For a specific subject's transcript $S$, we construct the co-occurrence graph $\mathcal{G}_{co}$ by mapping the pre-calculated normative weights onto the subject's actual word sequence. An edge $e_{ij} \in \mathcal{E}_{co}$ is established between word $w_i$ and any subsequent word $w_{i+k}$ (where $1 \leq k \leq n$) if and only if their normative PMI exceeds a predefined threshold $\tau_c$.


The structural density inherent in the co-occurrence graph $\mathcal{G}_{co}$ functions as a direct proxy for discourse coherence, effectively capturing the nuanced temporal "flow" of spontaneous speech. Within a healthy flow scenario—where a subject adheres to a logical and conventional narrative trajectory—the resulting $\mathcal{G}_{co}$ is defined by a high density of edges paired with significant PMI weights. Together, these elements construct a robustly well-connected topological "skeleton" composed of normative word associations. Conversely, the pathological disruption observed in patients with AD—which manifests as labored word retrieval or disjointed logical transitions—produces word pairs that are statistically rare or nonsensical compared to the normative baseline. This ultimately yields a fragmented graph topology marked by sparse or anomalous connectivity.

\subsection{View-Specific Encoding via GAT}
To learn the importance of different neighbors within each graph, we employ a Graph Attention Network (GAT) \cite{velickovic_graph_2018}. For a node $i$ in graph $k \in \{sem, syn, co\}$, the attention coefficient $\alpha_{ij}$ is computed as:
\begin{equation}
\alpha_{ij} = 
\frac{
  \exp\bigl(\operatorname{LeakyReLU}\bigl(\vec{a}^{\!\top} \bigl[ \mathbf{W}\mathbf{x}_i \concat \mathbf{W}\mathbf{x}_j \bigr]\bigr)\bigr)
}{
  \sum_{l \in \mathcal{N}_i} 
  \exp\bigl(\operatorname{LeakyReLU}\bigl(\vec{a}^{\!\top} \bigl[ \mathbf{W}\mathbf{x}_i \concat \mathbf{W}\mathbf{x}_l \bigr]\bigr)\bigr)
},
\end{equation}
where $\vec{a}$ and $\mathbf{W}$ are learnable parameters and $\mathcal{N}_i$ is the neighborhood of node $i$. The updated node features $\mathbf{h}'_i$ are aggregated to form a global graph representation $\mathbf{z}_k$ using Global Mean Pooling:
\begin{equation}
\mathbf{z}_k = \frac{1}{N} \sum_{i=1}^{N} \mathbf{h}'_i, \quad \mathbf{z}_k \in \mathbb{R}^{d_{gat}}.
\end{equation}

\subsection{Heterogeneity-Aware Gated Fusion}
Given that AD symptoms vary (e.g., some lose syntax, others lose semantic logic), we use a gated mechanism to dynamically weight the three views \cite{arevalo_gated_2017}.
First, we concatenate the three graph vectors: $\mathbf{Z}_{cat} = [\mathbf{z}_{sem} \concat \mathbf{z}_{syn} \concat \mathbf{z}_{co}]$. The gating network computes a weight vector $\mathbf{g}$:
\begin{equation}
\mathbf{g} = \text{Softmax}(\mathbf{W}_g \mathbf{Z}_{cat} + \mathbf{b}_g), \quad \mathbf{g} = [\beta_{sem}, \beta_{syn}, \beta_{co}].
\end{equation}
The fused representation $\mathbf{z}_{fused}$ is a weighted sum:
\begin{equation}
\mathbf{z}_{fused} = \beta_{sem}\mathbf{z}_{sem} + \beta_{syn}\mathbf{z}_{syn} + \beta_{co}\mathbf{z}_{co}.
\end{equation}
To prevent the loss of individual view details during fusion, we apply a pass-through concatenation:
\begin{equation}
\mathbf{z}_{\text{final}} = 
\bigl[ 
  \mathbf{z}_{\text{fused}} \concat 
  \mathbf{z}_{\text{sem}} \concat 
  \mathbf{z}_{\text{syn}} \concat 
  \mathbf{z}_{\text{co}} 
\bigr].
\end{equation}
This ensures that the classifier has access to both the "optimized mixture" and the raw specific features of each linguistic dimension.

\subsection{Classification and Objective Function}
The final vector $\mathbf{z}_{final}$ is passed through a Multi-Layer Perceptron (MLP). To classify the subject as either AD or Healthy Control (HC), the model outputs a predicted probability $\hat{y} = \text{Sigmoid}(\text{MLP}(\mathbf{z}_{final}))$.
To improve the model's generalization and prevent over-confident predictions, we employ Label Smoothing during training. Instead of using hard binary labels $y \in \{0, 1\}$, we transform them into soft targets $y_{\text{ls}}$:
\begin{equation}
y_{\text{ls}} = y(1 - \alpha) + \frac{\alpha}{K},
\end{equation}
where $\alpha = 0.2$ is the smoothing parameter and $K=2$ is the number of classes. The model is then optimized using the Smoothed Binary Cross-Entropy (BCE) Loss:
\begin{equation}
\mathcal{L} = -\frac{1}{M} \sum_{m=1}^{M} [y_{\text{ls}, m} \log(\hat{y}_m) + (1 - y_{\text{ls}, m}) \log(1 - \hat{y}_m)].
\end{equation}
This approach encourages the model to learn more robust latent representations by penalizing extreme logit values, thereby enhancing both calibration and classification performance on the clinical dataset.

\section{Experiments}
\subsection{Experimental settings}
\subsubsection{Dataset}
Our experiments were conducted using the standardized ADReSSo 2021 Challenge dataset \cite{luz_detecting_2021}, which is a curated subset of the Pitt Corpus within the DementiaBank database. Samples were collected via the "Cookie Theft" picture description task—a clinical gold standard for assessing narrative speech and cognitive status. The corpus comprises audio recordings and transcripts from 237 English speakers, including 122 individuals diagnosed with AD and 115 HC. Following the ADReSSo 2021 protocol, the data is partitioned into a training set of 166 participants (87 AD, 79 HC) and a test set of 71 participants (35 AD, 36 HC). We cropped the raw audio files based on provided timestamps to retain only segments containing subject speech. Since our approach constructs graph nodes directly from word-level tokens in subject narratives, samples lacking valid subject-specific segments—specifically five from the training set and one from the test set—were excluded as they could not support graph construction. To ensure experimental fairness and rigorous comparison, we re-implemented all baseline methods and evaluated them on this filtered version of the dataset.

\subsubsection{Implementation details}
All experiments were conducted on an NVIDIA GeForce RTX 4090D GPU. For the model architecture, the input dimension was set to 768 to match the BERT-base embeddings. The GAT component consisted of a single layer with 2 attention heads and a hidden dimension of 128. The subsequent MLP utilized a hidden layer of 256 units. To prevent overfitting, we applied a dropout rate of 0.5 and a weight decay of 0.003. Regarding the hyperparameter configuration, the model was trained for a maximum of 100 epochs with a batch size of 8. We employed the Adam optimizer with an initial learning rate of $1 \times 10^{-3}$, coupled with a ReduceLROnPlateau scheduler. Furthermore, we implemented label smoothing of 0.2 to improve the model's calibration and robustness. Through extensive experimentation, we determined the optimal hyperparameters for graph construction: a sliding window size of $n=3$ and an edge addition threshold of $\tau_c = 0.3$ for the co-occurrence graph, and a threshold of $\tau_s = 0.8$ for the semantic graph, which collectively yielded the best performance in our final results.

\subsection{Main results and discussions}
As illustrated in Table~\ref{tab:comparison}, our proposed framework demonstrates superior performance, consistently outperforming all baseline methods across all metrics. Specifically, it achieves a 5-fold cross-validation accuracy of 88.81\% on the training set and an accuracy of 90.00\% on the test set.

\begin{table*}[t] 
\centering
\caption{Performance comparison of different methods on 5-fold cross-validation and test set (\%).}
\label{tab:comparison}
\begin{tabular}{@{}lcccccccc@{}} 
\toprule
\multirow{2}{*}{\textbf{Method}} & \multicolumn{4}{c}{\textbf{5-Fold Cross-Validation Results}} & \multicolumn{4}{c}{\textbf{Test Set Results}} \\ 
\cmidrule(lr){2-5} \cmidrule(l){6-9}
 & \textbf{Accuracy} & \textbf{F1-Score}& \textbf{Recall} & \textbf{Precision}& \textbf{Accuracy} & \textbf{F1-Score}& \textbf{Recall} & \textbf{Precision}\\ 
\midrule
Luz et al. \cite{luz_detecting_2021}      & 78.22& 77.96& 73.40& 84.17& 79.71& 78.12& 73.53& 83.33\\
Balagopalan et al. \cite{balagopalan_bert_2020}& 80.78& 82.24& 83.79& 81.28& 82.86& 83.33& 85.71&81.08\\
 Ajroudi et al. \cite{ajroudi_exploring_2024}  & 85.08& 85.58& 83.86& 88.43& 81.43& 81.16& 80.00&82.35\\
Cai et al. \cite{cai_exploring_2023}      & 84.38& 85.54& 86.14& 85.10& 84.29 & 83.58& 80.00&87.50\\
Ortiz-Perez et al. \cite{ortiz-perez_cognialign_2025} & 86.88& 86.71& 87.31& 86.79& 81.43 & 80.60& 77.14 & 84.38\\
\midrule
\textbf{Ours}        & \textbf{88.81}& \textbf{88.81}& \textbf{88.81}& \textbf{89.43}& \textbf{90.00} & \textbf{89.86}& \textbf{88.57} & \textbf{91.18}\\
\bottomrule
\end{tabular}
\end{table*}

Compared to the baseline established by Luz et al. , our
model demonstrates a substantial performance leap. In contrast to Ajroudi et al., who utilized OpenAI’s text embeddings as high-dimensional features for direct classification, our approach employs BERT to establish a rigorous word-to-node mapping via token averaging. This strategy leverages deep semantic representations while simultaneously preserving the discrete symbolic structure of the text. Unlike the approach by Balagopalan et al., which utilizes BERT hidden features for direct classification, our multi-view gated graph attention network framework enables structural linguistic modeling by characterizing speech patterns through complementary perspectives.

While Cai et al.  focused on dependency or dynamic graphs, our integration of a PMI-based co-occurrence graph effectively captures temporal anomalies in speech flow. Pathological markers common in AD patients—such as disjointed logical transitions or labored word retrieval—are explicitly manifested through abnormal fluctuations in PMI weights. Simultaneously, the dependency graph models the integrity of the linguistic "skeleton," while the semantic graph utilizes cosine similarity to detect subtle semantic drift.

The method proposed by Ortiz-Perez et al. achieved state-of-the-art performance under the cross-validation setting. However, it was observed that this configuration leads to a significant performance degradation when the model is evaluated on the test set. This phenomenon may be attributed to the model overfitting to its respective validation split in each fold, yielding inflated overall metrics. For a more stringent assessment of performance, we evaluated the proposed approach against the test set, yielding an accuracy rate of 81.43\%.

\subsection{Ablation Study}
To evaluate the individual contribution of each architectural component to the diagnostic performance of our proposed method, we conducted a series of ablation experiments on the test set. By systematically removing specific graph views and the fusion mechanism, we generated five ablation variants for comparison. The results are summarized in Table~\ref{tab:ablation_study}.

The most significant performance degradation occurred in the w/o Graph Structure variant, which relies solely on the average pooling of BERT embeddings without any topological modeling. This variant achieved the lowest scores across all metrics (Acc: 81.43\%, F1: 80.60\%), underscoring the critical importance of capturing non-linear structural dependencies in pathological speech.

Among the specific linguistic views, the removal of the Co-occurrence Graph led to a substantial decline in the F1-score to 85.29\%. This confirms that the temporal "flow" and logical progression captured via PMI are indispensable biomarkers for Alzheimer's Disease. Similarly, the removal of the Semantic Graph and Dependency Graph resulted in accuracy drops to 85.71\% and 87.14\%, respectively. These results underscore that while semantic content provides the "substance" of the diagnosis, the syntactic "skeleton" provides the necessary linguistic constraints for robust classification.

The w/o Gated Fusion variant, which utilizes a static averaging of the three graph outputs instead of the dynamic weighting network, saw its accuracy fall to 87.14\%. This confirms that a "one-size-fits-all" approach to linguistic feature integration is insufficient for AD detection. The gated network’s ability to adaptively prioritize different linguistic views based on individual patient speech patterns—such as prioritizing syntactic errors in one patient and semantic incoherence in another—is a critical factor in achieving the full model's accuracy of 90.00\%.

\begin{table}[t]
\centering
\caption{Ablation study results on test set (\%).}
\label{tab:ablation_study}
\begin{tabular}{lcccc}
\toprule
\textbf{Model Variant} & \textbf{Accuracy} & \textbf{F1-Score} \\
\midrule
w/o Semantic Graph & 85.71 & 86.11 \\
w/o Dependency Graph & 87.14 & 86.57 \\
w/o Co-occurrence Graph & 85.71 & 85.29 \\
w/o Gated Fusion & 87.14 & 87.32 \\
w/o Graph Structure & 81.43 & 80.60 \\
\midrule
\textbf{Ours} & \textbf{90.00} & \textbf{89.86} \\
\bottomrule
\end{tabular}
\end{table}

\section{Conclusion}
Experimental evaluations on the ADReSSo 2021 dataset validate that the Multi-View Gated Graph Attention Network provides a robust framework for identifying AD by holistically characterizing the "content-structure-flow" features of spontaneous speech. One of the primary contributions of this study is the pivotal role played by the co-occurrence graph; by utilizing normative PMI to quantify discourse logic, it successfully captures the incoherent logical transitions and repetitive patterns characteristic of cognitive decline that traditional sequential models are typically unable to detect. Ablation results indicate that removing this "flow" dimension significantly impairs diagnostic performance, confirming that narrative progression is a crucial feature. Furthermore, unlike conventional static fusion strategies, our gated network adaptively assigns weights to linguistic features on a per-sample basis, effectively mitigating the challenges posed by the inherent clinical heterogeneity of AD. Future research will prioritize evolving the framework into a comprehensive multimodal system by integrating acoustic prosody as additional graph-based perspectives.

\section{Acknowledgments}
This work was supported by the National Natural Science Foundation of China under Grant U23B2053 and by the National Talent Program under Grants E4G008, E55304, E43301, and E476.

\section{Generative AI Use Disclosure}
During the preparation of this manuscript, we used generative AI tools to polish the English language and improve readability. These tools were not used to generate any scientific claims, experimental results, or significant parts of the manuscript.

\bibliographystyle{IEEEtran}
\bibliography{reference}

\end{document}